\documentclass[
]{ceurart}

\usepackage{listings}
\usepackage{longtable}
\usepackage{booktabs}
\begin{document}

%%
%% Rights management information.
%% CC-BY is default license.
\copyrightyear{2022}
\copyrightclause{Copyright for this paper by its authors.
  Use permitted under Creative Commons License Attribution 4.0
  International (CC BY 4.0).}

%%
%% This command is for the conference information
\conference{De-Factify 4: 4th Workshop on Multimodal Fact Checking and Hate Speech Detection, co-located with AAAI 2025. Pennsylvania.}

%%
%% The "title" command
\title{OSINT at CT2 - AI-Generated Text Detection: Tracing Thought: Using Chain-of-Thought Reasoning to Identify the LLM Behind AI-Generated Text}

\tnotemark[1]
\tnotetext[1]{You can use this document as the template for preparing your
  publication. We recommend using the latest version of the ceurart style.}

%%
%% The "author" command and its associated commands are used to define
%% the authors and their affiliations.
\author[1]{Shifali Agrahari }[%
email=a.shifali@iitg.ac.in
]
\cormark[1]
\fnmark[1]
\address[1]{Department of Computer Science and Engineering Indian Institute of Technology Guwahati, India}

\author[1]{Sanasam Ranbir Singh}[%
email= ranbir@iitg.ac.in
]
\fnmark[1]

%% Footnotes
\cortext[1]{Corresponding author.}
\fntext[1]{These authors contributed equally.}

%%
%% The abstract is a short summary of the work to be presented in the
%% article.
\begin{abstract}
In recent years, the detection of AI-generated text has become a critical area of research due to concerns about academic integrity, misinformation, and ethical AI deployment. This paper presents \texttt{COT\_Finetuned}, a novel framework for detecting AI-generated text and identifying the specific language model (LLM) responsible for generating the text. We propose a dual-task approach, where Task A involves classifying text as AI-generated or human-written, and Task B identifies the specific LLM behind the text. The key innovation of our method lies in the use of Chain-of-Thought (CoT) reasoning, which enables the model to generate explanations for its predictions, enhancing transparency and interpretability. Our experiments demonstrate that \texttt{COT\_Finetuned} achieves high accuracy in both tasks, with strong performance in LLM identification and human-AI classification. We also show that the CoT reasoning process contributes significantly to the model’s effectiveness and interpretability.

\end{abstract}

%%
%% Keywords. The author(s) should pick words that accurately describe
%% the work being presented. Separate the keywords with commas.

%%
%% This command processes the author and affiliation and title
%% information and builds the first part of the formatted document.
\maketitle

\section{Introduction}
Recent advancements in Natural Language Processing (NLP) have led to the development of powerful pre-trained language models (PLMs) capable of generating highly realistic text. These models, such as GPT-4, DeBERTa, and T5, have made significant strides in a wide range of applications, from chatbots and text generation to machine translation and summarization. However, as the capabilities of these models increase, so does the challenge of distinguishing between human-generated and AI-generated text. This has raised concerns about academic integrity, misinformation, and responsible deployment of AI technologies.

To address these issues, The AAAI 2025 DEFACTIFY 4.0 \cite{roy-2025-defactify-overview-text} – Workshop Series on Multimodal Fact-Checking and Hate Speech Detection on CT2 - AI Generated Text Detection Task A: This is a binary classification task where the goal is to determine whether each given text document was generated by AI or created by a human.
Task B: Building on Task A, this task requires participants to identify which specific LLM generated a given piece of AI-generated text. For this task, participants will know that the text is AI-generated and must predict whether it was produced by models such as GPT 4.0, DeBERTa, FalconMamba, Phi-3.5, or others.

In this paper, we propose a novel framework, Chain of Thought Finetuning (\texttt{COT\_Finetuned}), for the dual task problem of AI-generated text detection and identification of the specific language model (LLM) responsible for the generation of the text. Our approach builds upon the concept of Chain-of-Thought (CoT) reasoning, which provides a structured and explainable process for classifying text and identifying the model behind its generation. 

To solve these tasks, we introduce a fine-tuned model that is capable of jointly predicting both the AI vs. Human classification (Task A) and the specific LLM responsible for generating the text (Task B). The key innovation in our approach is the use of Chain-of-Thought (CoT) reasoning, which allows the model to generate explanations for its decisions, enhancing interpretability and transparency. These explanations not only provide insights into the classification process but also assist in understanding the stylistic choices and patterns unique to different LLMs.

 \begin{figure*}[t]
    \centering
    \includegraphics[width=
\textwidth]{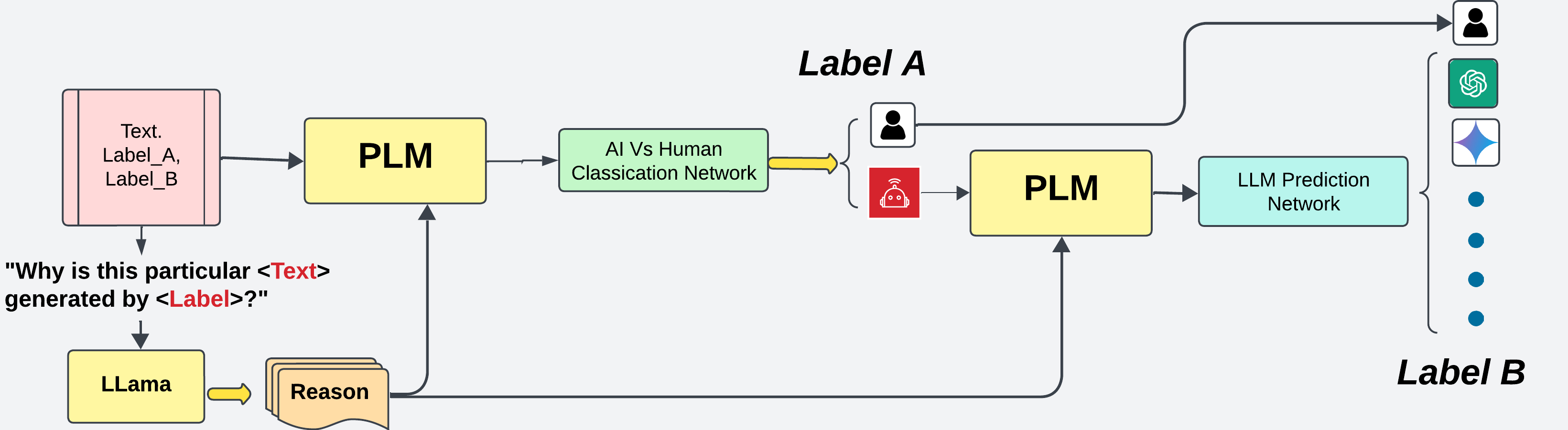} % Replace with the path to your imag
    \caption{Proposed detector model for binary classification task A \& multi classification task B.}
    \label{fig:prop}
\end{figure*}
\section{Related Work}

Detecting AI-generated text has gained significant attention in recent years, with numerous methods proposed to tackle this challenge. Typically, this task is framed as a binary classification problem, distinguishing human-written text from machine-generated content \cite{zellers2019defending, gehrmann2019gltr, ippolito2019automatic}. Existing approaches can be broadly categorized into three main types: supervised methods, unsupervised methods, and hybrid approaches.

Supervised methods rely on labeled datasets to train detection models. Research in this area includes works like \cite{Wang2023ImplementingBA, uchendu2021turingbench, zellers2019defending, zhong2020neural, liu2023argugpt, liu2022coco}, which demonstrate that supervised approaches generally achieve strong performance. However, these methods are prone to overfitting, particularly when dealing with limited or domain-specific training data \cite{mitchell2023detectgpt, su2023detectllm}. Unsupervised methods, such as zero-shot detection techniques \cite{solaiman2019release, ippolito2019automatic, mitchell2023detectgpt, su2023detectllm, hans2401spotting, shijaku2023chatgpt}, leverage the capabilities of pre-trained language models to classify text without task-specific training. Adversarial methods also fall within this category, focusing on evaluating detector robustness against perturbations. For example, \cite{antoun2023towards} assesses the impact of character-level modifications like misspellings, using French as a case study. Similarly, \cite{krishna2024paraphrasing} introduces DIPPER, a generative model trained to paraphrase paragraphs and evade detection.

Hybrid approaches combine feature-based methods with machine learning or neural models. They often utilize metrics such as word count, vocabulary richness, and readability scores, fused with machine learning or fine-tuned neural networks for detection \cite{solaiman2019release, shah2023detecting, nguyen2017identifying, mindner2023classification, kumarage2023neural}. Fusion and ensemble strategies have also been explored to enhance detection accuracy.

\section{Methodology}

\subsection{Overview}
In this paper, we propose a novel method called \texttt{COT\_Finetuned} that combines chain-of-thought (CoT) reasoning with fine-tuned classification models to identify whether a given text document is AI-generated or human-written, and to further identify which specific Language Model (LLM) generated an AI-generated document. The method is designed to handle two tasks simultaneously:

\begin{itemize}
    \item \textbf{Task A:} Binary classification to determine if the text is AI-generated or human-written.
    \item \textbf{Task B:} For AI-generated texts, multi-class classification to identify which LLM (e.g., GPT-4.0, DeBERTa, FalconMamba, Phi-3.5, etc.) generated the text.
\end{itemize}

\subsection{COT\_Finetuned Method}
The \texttt{COT\_Finetuned} method uses a chain-of-thought reasoning process to explain and classify the given text documents. The method generates two outputs: a classification result for Task A and, if applicable, an identification of the LLM for Task B, along with the reasoning behind each classification.

\subsubsection{Dataset}
The dataset consists of text documents \( d_i \), each paired with two labels for classification:
\begin{itemize}
    \item \( \text{Label}_A \in \{0, 1\} \): The label for Task A, where 0 indicates that the document is human-written and 1 indicates the document is AI-generated.
    \item \( \text{Label}_B \in \{ \text{GPT-4.0}, \text{DeBERTa}, \text{FalconMamba}, \text{Phi-3.5}, \dots \} \): The label for Task B, indicating the specific LLM that generated the text (if \( \text{Label}_A = 1 \)).
\end{itemize}

In addition, the model will generate the reasoning for its classification decision based on the provided labels.

% \subsubsection{Method Process}
% The method operates in the following manner:

% \begin{itemize}
%     \item \textbf{Initial Classification (Task A)}: The model first classifies the text as either human-written or AI-generated. If \( \text{Label}_A = 0 \), the document is classified as human-written.
%     \item \textbf{LLM Identification (Task B)}: If the document is AI-generated (\( \text{Label}_A = 1 \)), the model then identifies the specific LLM used to generate the text. This is done by analyzing the text's structure, style, and vocabulary, and comparing them with known patterns from various LLMs.
%     \item \textbf{Chain-of-Thought Reasoning Generation}: The model generates a reasoning prompt to justify its decision-making. For each document, the model is prompted to answer the following question:
    
%     \begin{quote}
%         \textit{Why is this particular "Text" generated by "Label"?}
%     \end{quote}
    
%     The prompt asks the model to explain why the text fits the classification and LLM identification by analyzing stylistic and linguistic features.
% \end{itemize}
\subsubsection{Method Process}
Let \( d_i \) be the document, and let \( \text{Label}_A \) and \( \text{Label}_B \) represent the labels for Task A and Task B, respectively.

For Task A (binary classification) \& Task B (multi-class classification):
\begin{equation}
\text{If } \text{Label}_A = 0, \quad \text{then } \text{Label}_B = \text{Human}.
\text{If } \text{Label}_A = 1, \quad \text{then }  \hat{\text{Label}}_B = \arg\max_j \mathcal{L}(\mathcal{M}_j \mid d_i),
\end{equation}
where \( \mathcal{L}(\mathcal{M}_j \mid d_i) \) is the likelihood of model \( \mathcal{M}_j \) generating the document \( d_i \). The model assigns \( \text{Label}_B \) based on the highest likelihood.

\subsubsection{Loss Function}
The model is fine-tuned using a combined loss function that incorporates the classification loss for both Task A and Task B, along with a reasoning loss component.

\begin{itemize}
    \item \textbf{Classification Loss}: Binary cross-entropy loss is used for Task A (human-written vs. AI-generated), while cross-entropy loss is employed for Task B (identifying the specific LLM for AI-generated documents):
    \begin{equation}
    \mathcal{L}_{\text{classification}}(\hat{y}_i, y_i) = - \left( y_i \cdot \log(\hat{y}_i) + (1 - y_i) \cdot \log(1 - \hat{y}_i) \right).
    \end{equation}

 %   \item \textbf{Reasoning Loss}: This quantifies the quality of the generated reasoning by comparing the generated reasoning \( r_i \) with the true reasoning \( r_i^{\text{true}} \), using a similarity metric such as BLEU or ROUGE:
    %\begin{equation}
  %  \mathcal{L}_{\text{reasoning}}(r_i, r_i^{\text{true}}) = \text{Similarity}(r_i, r_i^{\text{true}}).
  %  \end{equation}

    \item \textbf{Total Loss}: The total loss combines the both classification task:
    \begin{equation}
    \mathcal{L}_{\text{total}} = \mathcal{L}_{\text{classification}}(\hat{y}_i^A, y_i^A) + \mathcal{L}_{\text{classification}}(\hat{y}_i^B, y_i^B) 
    \end{equation}
\end{itemize}

\subsubsection{Training and Fine-Tuning}
The model is fine-tuned on a training dataset consisting of text documents \( d_i \), the true labels for Task A (\( y_i \)) and Task B (\( y_i^B \)), and their corresponding reasoning (\( r_i \)). The training process involves optimizing the model parameters to minimize the total loss using gradient-based optimization algorithms such as Adam.

\textbf{Process Overview}:
\begin{enumerate}
    \item Each document \( d_i \) is labeled as either AI-generated or human-written (\( y_i \in \{\text{AI}, \text{Human}\} \)).
\item For each document \( d_i \), a prompt \( p(d_i, y_i) \) is created:
\begin{equation}
p(d_i, y_i) = \text{"Why is this particular \( d_i \) generated by \( y_i \)?"}.    
\end{equation}

\item The prompt is passed to the LLaMA model, generating a reasoning \( r_i \):
\begin{equation}
o_i = \text{LLaMA}(p(d_i, y_i)) \rightarrow r_i,  
\end{equation}

where \( r_i \) provides insights into why \( d_i \) is classified as either AI-generated or human-written.
\item For binary classification (Task A), the text \( d_i \), label \( y_i \), and reasoning \( r_i \) are passed to a pre-trained language model (PLM) such as BERT. The output layer applies a sigmoid activation for binary classification.

\item If \( \text{Label}_A = 1 \) (AI-generated), the future text \( d_i \), label \( y_i^B \), and reasoning \( r_i^B \) are passed to a fine-tuned PLM for multi-class classification of Task B.
\end{enumerate}
By iteratively optimizing the loss, the model learns to classify documents and generate high-quality reasoning, ensuring accurate and interpretable predictions. After completing the training, pass the test dataset to the model, which will return Labe\_A and Label\_B.

\section{Experiments}
\subsection{Dataset and Processioning}
% \begin{document}

For each task, the organizers provided three datasets \cite{roy-2025-defactify-dataset-text}: Train, Dev and Test snap mentioned in appendix. Table \ref{dataset}. Training and development data with labels (AI or human) for the development phase and for the evaluation phase, testing data without labels for both tasks. All descriptions with respect to the size of dataset is mentioned in Table~\ref{tab:detail_dataset}.

\begin{table}[ht]
\centering
\begin{tabular}{|l|l|l|l|l|}
\hline
\textbf{Train set} & \textbf{Test set} & \textbf{Val set} & \textbf{Total} \\ \hline
7320 & 1570 & 1570 & 10500 \\ \hline
\end{tabular}
\label{tab:detail_dataset}
\caption{Statistics of the dataset used in the study, showing the number of samples in the training, test, and validation sets.}
\end{table}
The goal of this task is to classify each text document as AI-generated or human-written. Let the data set consist of text documents \( D = \{ d_1, d_2, \dots, d_n \} \), where each document \( d_i \) is labeled as AI-generated or human-written. Specifically, for each document \( d_i \), a label \( y_i \in \{ \text{AI}, \text{Human} \} \) is assigned, where \( y_i = \text{Human} \) indicates a human-written document and \( y_i = \text{AI} \) indicates an AI-generated document.
The training dataset consists of tuples \( \{ (d_i, y_i, y_i^B, r_i, r_i^B) \} \), where: \( d_i \): The input text document, \( y_i \): The true label for Task A (AI-generated or Human-written), \( y_i^B \): The true label for Task B (specific LLM if AI-generated) and \( r_i \), \( r_i^B \): The reasoning behind the classification decision.

\subsection{Experimental Setup}

For both Task, the hyperparameters include an epoch size ranging from 50 to 250, while the batch size is fixed at 32, determined by the available GPU resources. Further details of the experimental setup are presented in the Appendix \ref{dataset}. For this experiment, we consider pre-trained language models such as \textit{BERT} for both tasks.

\section{Results and Analysis}
% In this section, we analyze the results of Subtask A and Subtask B. S
% ubtask A focused on feature extraction and evaluating different models, while Subtask B also involved feature extraction with a different set of models. Both subtasks aimed to assess the increase the classwise F1 score to evaluate model performance.

%\subsection{Subtask A}

Table \ref{task_result} show the results of the Leaderboard on test dataset. Table 3 shows the F1-scores for different methods applied to Task A (binary classification) and Task B (multi-class classification). 

For Task A, Bert outperforms Roberta with an F1-score of 0.742 compared to 0.672. The addition of Chain-of-Thought (COT) reasoning improves the performance of both models, with Bert + COT achieving the highest F1-score of 0.898. 

For Task B, the F1-scores are lower across all methods, reflecting the increased complexity of the multi-class classification task. Bert + COT outperforms all other methods with a score of 0.307, while Roberta + COT achieves 0.198. 

Overall, COT reasoning improves performance in both tasks, with Bert + COT being the most effective method, especially for Task A. However, Task B remains challenging, highlighting the need for further improvements in multi-class classification.

% Our main configuration achieved strong results in both English and Arabic tasks as mentioned in Table \ref{tab:performance_metrics}, which is shown in the leader board of this task with the rank of our model.
%Finally, For Subtask A: Our proposed model \textit{DeBERTa-base} + faetures achieves the highest score of 0.978 and for Subtask B: Our proposed model \textit{AraBERT} + faetures  outperforms all models with an F1 score of 0.9429.Although for Subtask A, other model also perform better with features and For Subtask B, Arabert without features achevies 0.9214 because Araberta model specially designed by Arabic Language to it extract all the important features. Fig. \ref{dev_sult}, confusion matrics for Deveploment dataset on Our proposed model for both Subtask respectively. Arabic data are more misclassified as compare to English.

\begin{table}[ht]
\centering
\begin{tabular}{|c|l|l|c|c|}
\hline
\textbf{S.No} & \textbf{Name} & \textbf{Team Name} & \textbf{Score for Task-A} & \textbf{Score for Task-B} \\ \hline
1 & Avinash Trivedi & Sarang & 1 & 0.9531 \\ \hline
2 & Duong Anh Kiet & dakiet & 0.9999 & 0.9082 \\ \hline
3 & Vijayasaradhi Indurthi & tesla & 0.9962 & 0.9218 \\ \hline
4 & Shrikant Malviya & SKDU & 0.9945 & 0.7615 \\ \hline
5 & Harika Abburi & Drocks & 0.9941 & 0.627 \\ \hline
6 & Manoj Saravanan & Llama\_Mamba & 0.988 & 0.4551 \\ \hline
7 & Chinnappa Guggilla & AI\_Blues & 0.9547 & 0.4698 \\ \hline
8 & Xinlong Zhang & NLP\_great & 0.9157 & 0.1874 \\ \hline
\textbf{9} & \textbf{Shifali Agrahari} & \textbf{Osint} & \textbf{0.8982 }& \textbf{0.3072} \\ \hline
10 & Xiaoyu & Xiaoyu & 0.803 & 0.5696 \\ \hline
11 & Rohan R & Rohan & 0.7546 & 0.4053 \\ \hline
\end{tabular}
\label{task_result}
\caption{Presents the final leaderboard scores for all participating teams in Task A and Task B. The scores have been officially provided by the organizer, reflecting the performance of each team based on the F1 score.}
\end{table}

\begin{table}[ht]
    \centering
    \begin{tabular}{|l|c|c|}
    \hline
        \textbf{Method} & \textbf{Score for Task-A} & \textbf{Score for Task-B}   \\
 
 \hline
        RoBERTa & 0.672 & 0.143  \\ 
     BERT  & 0.742 & 0.249   \\  
     RoBERTa + COT  & 0.792 & 0.198   \\ 
     \textbf{BERT + COT}  & 0.898 & 0.307   \\ 
      \hline
    \end{tabular}

    \caption{F1-Scores of Different Methods for Task-A and Task-B}
    \label{tab:performance_metrics}
\end{table}
\vspace{-0.5cm}
\section{Conclusion}

In this paper, we introduced \texttt{COT\_Finetuned}, a dual-task framework for detecting AI-generated text and identifying the specific language model (LLM) that produced it. By leveraging Chain-of-Thought (CoT) reasoning, we not only achieved strong performance in distinguishing between human and AI-generated text but also provided explanations for the model’s decisions, contributing to the transparency and interpretability of AI classification tasks. Our experiments on a real-world dataset showed that \texttt{COT\_Finetuned} performs competitively in both tasks, outperforming traditional classification models while offering valuable insights into the text generation process.

We also observed that the CoT reasoning mechanism enhances the model’s understanding of the text’s stylistic features, enabling it to better identify subtle patterns associated with different LLMs. The dual-task nature of the framework allows for a more comprehensive analysis of AI-generated content, making it useful for applications in academic integrity, content moderation, and AI ethics.

In conclusion, the proposed method offers an effective solution to the challenges of detecting AI-generated text and identifying the underlying models, providing both high accuracy and interpretability.

\bibliography{main}

%%
%% If your work has an appendix, this is the place to put it.
\appendix
\section{Appendix}
\subsection{Dataset detail}
\label{dataset}
\begin{table}[ht]
\centering
\begin{tabular}{|p{1.6cm}|p{1.5cm}|p{1.8cm}|p{1.5cm}|p{1.8cm}|p{1.7cm}|p{1.6cm}|p{1.8cm}|}
\hline
\textbf{Prompt} & \textbf{Human} & \textbf{Gemma-2-9B} & \textbf{Mistral-7B} & \textbf{Qwen-2-72B} & \textbf{Llama-8B} & \textbf{Yi-Large} & \textbf{GPT-4-o} \\ \hline
Roberta Karmel, First Woman Named to the S.E.C., Dies at 86. & Roberta Karmel, the first woman ........... & Roberta Karmel, who made history as the first woman appointed ........ & Roberta Karmel, a trailblazer who became...... & Roberta Karmel, Securities and Exchange Commission........... & Roberta Karmel, the first female commissioner of....... & Roberta Karmel, the first woman to serve on the U.S6....... & Roberta Karmel, who made history first woman appointed to the S.E.C........ \\ \hline
In the age of coronavirus, the only way you can see Milan is to fly through it. & Messages From Quarantine transcript  .......... & \# In the Age of Coronavirus, the Only Way You Can See Milan is to Fly....... & Title: "Exploring Milan in the Age of Coronavirus: .......... & In the Age of Coronavirus, the Only Way to See Milan is to Fly ....... & I**The New York Times** **IN THE AGE OF CORONAVIRUS,....... & **Title: Navigating Milan in the Age of Coronavirus: ....... & **Title: In the Age of Coronavirus, the Only Way *......... \\ \hline

\end{tabular}
\label{tab:dataset}
\caption{Comparison of Text Generated by Different Models}
\end{table}

\begin{table}[ht]
\centering
\begin{tabular}{|p{6cm}|c|c|}
\hline
\textbf{Text} & \textbf{Label\_A} & \textbf{Label\_B} \\ \hline
Roberta Karmel, the first woman appointed to the U.S. Securities and Exchange Commission (S.E.C.), passed away at 86........... & 0 & Human \\ \hline
 \# In the Age of Coronavirus, the Only Way You Can See Milan is to Fly Through It Milan, the fashion capital of the world, is known for its bustling piazzas....... & 1 & Gemma-2-
9B \\ \hline
 Roberta Karmel, a trailblazer who became the first woman to serve on the S.E.C., has
died at the
age of ..... & 1 & Mistral-7B \\ \hline

\end{tabular}
\label{tab:dataset}
\caption{Span of Training Dataset}
\end{table}
% \section{Online Resources}

\subsection{Training detail}
\begin{table}[h!]
\centering
\begin{tabular}{|l|p{4.5cm}|}
\hline
\textbf{Hyperparameter} & \textbf{Setup: Fine-tuning PLM} \\ \hline
Epochs                  & 10-250                             \\ \hline
Batch Size              & 5                                \\ \hline
k            & 6 layer                                \\ \hline
Learning Rate           & \(2 \times 10^{-5}\)             \\ \hline
Optimizer               & Adam                              \\ \hline
L2 Regularization       & Weight decay: 0.01               \\ \hline
Loss Function           &      Classification \& Reasoning Loss         \\ \hline
\end{tabular}
\caption{Hyperparameter settings for Setup 1: Fine-tuning PLM.}
\label{tab:hyperparameters_setup1}
\end{table}

\begin{longtable}{p{0.3\linewidth} p{0.15\linewidth} p{0.3\linewidth}}
    \caption{Classification and Reasoning of Review Texts} \\
    \toprule
    \textbf{Review Text} & \textbf{Classification} & \textbf{Reasoning} \\
    \midrule
    \endfirsthead

    \toprule
    \textbf{Review Text} & \textbf{Classification} & \textbf{Reasoning} \\
    \midrule
    \endhead

    \midrule
    \multicolumn{3}{r}{\textit{Continued on next page}} \\
    \endfoot

    \bottomrule
    \endlastfoot

    I just received this dress and I'm blown away by how well it fits!  
    & Human 
    & Personal excitement and specific details about the fit and quality. Emotional tone and specific product-related commentary suggest a human review. \\
\\ \hline
    Theft room hotel staff stayed hotel beginning September, evening checking returned money stolen room. Our room door wasn't forced entry staff, reported girl 
    & Human 
    & Describes a negative experience with a lot of detail about a theft incident, hotel response, and follow-up efforts. The writing style has imperfections, a hallmark of a human review. \\

 \\ \hline
I just wore this to a wedding and I'm absolutely obsessed! It's the most flattering dress I've ever owned. The material is so soft and drapes perfectly, and it's incredibly comfortable... 
    Silent fresh moment 
    & LLM 
    & Very brief and lacks specific detail or personal involvement. Feels like a generated response without the emotional depth or depth of thought typical in human reviews. \\

\end{longtable}

% The sources for the ceur-art style are available via
% \begin{itemize}
% \item \href{https://github.com/yamadharma/ceurart}{GitHub},
% % \item \href{https://www.overleaf.com/project/5e76702c4acae70001d3bc87}{Overleaf},
% \item
%   \href{https://www.overleaf.com/latex/templates/template-for-submissions-to-ceur-workshop-proceedings-ceur-ws-dot-org/pkfscdkgkhcq}{Overleaf
%     template}.
% \end{itemize}

\end{document}